\documentclass{article}

\usepackage[english]{babel}

\usepackage[a4paper,top=2cm,bottom=2cm,left=3cm,right=3cm,marginparwidth=1.75cm]{geometry}
\usepackage{float}
\usepackage{amsmath}
\usepackage{graphicx}
\usepackage[colorlinks=true, allcolors=blue]{hyperref}
\usepackage{biblatex}
\usepackage{isomath}
\usepackage{amssymb}
\usepackage{algorithm}
\usepackage{algpseudocode}
\usepackage{comment}
\usepackage{booktabs}
\usepackage{tabularx}
\usepackage{multirow}
\usepackage{adjustbox}

\usepackage{caption}
\usepackage{setspace}
\usepackage{titlesec}
\usepackage{lipsum} 

\titlespacing*{\section}{0pt}{*1}{*1}
\titlespacing*{\subsection}{0pt}{*0.5}{*0.5}
\titlespacing*{\subsubsection}{0pt}{*0.5}{*0.5}

\newcommand{\fullfootcite}[1]{%
  \footnote{\fullcite{#1}}%
}
\title{Compare different SG-Schemes based on large least square problems}
\author{
    Ramkrishna Acharya\textsuperscript{1} \quad \texttt{qramkrishna@gamil.com} 
    \and 
    Durga Pokharel\textsuperscript{2} \quad \texttt{pokhareldurga88@gmail.com}
}
\date{
    \textsuperscript{1} FAU Erlangen-Nuremberg, Erlangen, Germany \\
    \textsuperscript{2} Herald College Kathmandu, Kathmandu, Nepal
}

\begin{document}

\maketitle

\begin{abstract}
This study reviews popular stochastic gradient-based schemes based on large least-square problems. These schemes, often called optimizers in machine learning, play a crucial role in finding better model parameters. Hence, this study focuses on viewing such optimizers with different hyper-parameters and analyzing them based on least square problems. Codes that produced results in this work are available on \href{https://github.com/q-viper/gradients-based-methods-on-large-least-square}{GitHub}.
\end{abstract}

\section{Introduction}
\subsection{Least Squares Problems}
Least squares problems are common problems in which we try to find the best-fitting curve to a given set of points by minimizing the sum of the squares of the offsets ("the residuals") of the points from the curve. \footnotetext[1]{\fullcite{MathWorld-LeastSquaresFitting}} Alternatively, they can be defined as a parameter estimation method in regression analysis based on minimizing the sum of squared residuals (RSS).

\vspace{1em}

Let:
\begin{itemize}
    \item $\mathbf{y} \in \mathbb{R}^n$  be the vector of observed values
    \item  $\mathbf{X} \in \mathbb{R}^{n \times p}$ be the matrix of input features
    \item $\boldsymbol{\theta} \in \mathbb{R}^p$ be the vector of regression coefficients (or parameters)
\end{itemize}

Then such a least squares problem can be defined as the matrix form of:

\begin{equation}
    \mathbf{y} = \mathbf{X} \boldsymbol{\theta} + \boldsymbol{\epsilon}
\end{equation}

where \( \boldsymbol{\epsilon} \in \mathbb{R}^n \) represents the error terms, typically assumed to be normally distributed.

Then the sum of squared residuals (RSS) can be written as:
\begin{equation}
    \text{RSS} = (\mathbf{y} - \mathbf{X} \boldsymbol{\theta})^T (\mathbf{y} - \mathbf{X} \boldsymbol{\theta})
\end{equation}

Alternatively, in terms of the error vector \( \boldsymbol{\epsilon} \), RSS can be expressed as:

\begin{equation}
    \text{RSS} = \boldsymbol{\epsilon}^T \boldsymbol{\epsilon} \notag
\end{equation}

\begin{equation}
    \text{RSS} = \boldsymbol{\epsilon}^T \boldsymbol{\epsilon} := (\mathbf{y} - \mathbf{X} \boldsymbol{\theta})^T (\mathbf{y} - \mathbf{X} \boldsymbol{\theta})
\end{equation}

The parameters $\theta$ can be found by minimizing the sum of squared residuals. Expanding on the above expression:
        \begin{equation}
            \boldsymbol{\epsilon}^T \boldsymbol{\epsilon} = \mathbf{y}^T \mathbf{y} - 2 \boldsymbol{\theta}^T \mathbf{X}^T \mathbf{y} + \boldsymbol{\theta}^T \mathbf{X}^T \mathbf{X} \boldsymbol{\theta} \notag
        \end{equation}
        To minimize RSS, we take the derivative with respect to \( \boldsymbol{\theta} \) and set it to zero:
        \begin{equation}
            \frac{\partial \boldsymbol{\epsilon}^T \boldsymbol{\epsilon}}{\partial \boldsymbol{\theta}} = -2 \mathbf{X}^T \mathbf{y} + 2 \mathbf{X}^T \mathbf{X} \boldsymbol{\theta} = 0 \notag
        \end{equation}
        Solving for \( \boldsymbol{\theta} \):
        \begin{equation}\label{eq:param1}
            \boldsymbol{\theta} = (\mathbf{X}^T \mathbf{X})^{-1} \mathbf{X}^T \mathbf{y}
        \end{equation}
        where \( \mathbf{X}^T \mathbf{X} \) is assumed to be invertible.

\section{Methodology}
\subsection {Least Squares and Gradient Scheme}
Gradient descent is an optimization algorithm used to minimize some convex functions by iteratively moving in the direction of the steepest descent as defined by the negative of the gradient. The parameter update technique is given by \ref{eq:param1} i.e. \(\boldsymbol{\hat{\theta}} = (\mathbf{X}^T \mathbf{X})^{-1} \mathbf{X}^T \mathbf{y}\) could have problems:

\begin{itemize}
    \item \((\mathbf{X}^T \mathbf{X})^{-1}\) might not be invertible.
    \item Our \( \mathbf{X} \) might be too big and cause storage problems while loading it.
    \item The number of observations (rows in \( \mathbf{X} \)) could be smaller than features (columns in \( \mathbf{X} \)).
\end{itemize}

But, we could find the best parameters \( \boldsymbol{\hat{\theta}} \) with an iterative update method. Let's define \( \mathbf{f}(\mathbf{X}; \boldsymbol{\theta}) = \mathbf{X}\boldsymbol{\theta} + \boldsymbol{\epsilon} \) as a regression function, and \( J(\mathbf{f}(\mathbf{X}; \boldsymbol{\theta}), \mathbf{y}) \) be a convex loss function we try to minimize. Then an example of such an iterative method can be shown in the algorithm \ref{alg:simple_gd}.

\begin{algorithm}[H]
\caption{Simple Gradient Descent Algorithm for Least Squares Regression}
\label{alg:simple_gd}
\begin{algorithmic}[1]
    \State Initialize parameters \( \boldsymbol{\theta} \), learning rate (step length) \(\eta\)
    \While{stopping criterion not met}
        \For{each data point \( (\mathbf{X}^{(i)}, y^{(i)}) \) in \( (\mathbf{X}, \mathbf{y}) \)}
            \State Compute gradient estimate: \( g= \frac{\partial}{\partial \theta} J(f(X^{(i)};\theta), y^{(i)}) \)
            \State Update parameters: \( \boldsymbol{\theta} \leftarrow \boldsymbol{\theta} - \eta g \)
        \EndFor
    \EndWhile
\end{algorithmic}
\end{algorithm}

\subsection{Different Choices to Make While Updating Parameters}

In machine learning, we use gradient descent to update the parameters of our model. In our case, we try to fit a linear regression model and try to find the best parameters $\theta$. The following section explains some of the different choices one can make:

\begin{itemize}
    \item Loss Functions : Example: Mean Squared Error (MSE), Mean Absolute Error (MAE), etc.
    \item Batch Size : Example: Batch size of 1, batch size of data counts, and any real values in between.
    \item Optimizers : Example: Variable learning rate, constant learning rate, momentum, etc.
\end{itemize}

\subsubsection{ Loss Functions}
Loss functions are the convex functions we try to minimize. By finding the gradients with respect to the parameters, we find the direction and magnitude needed for our parameter to be updated. Hence it plays a crucial role in gradient schemes. Mean Squared Error and Mean Absolute Error are the most common examples of convex functions.

\vspace{1em}

\textbf{1. Mean Squared Error (MSE)}
 \[
            \text{J}(\boldsymbol{\theta}) = \frac{1}{N} \sum_{i=1}^{N} (y^{(i)} - f(X^{(i)};\theta))^2 := \frac{1}{N} \sum_{i=1}^{N} j_i(\theta)
            \]
            where \( y^{(i)} \) are the observed values, and \( N \) is the number of data points. MSE is smooth and convex in nature and a good choice for regression tasks.

\vspace{1em}

\textbf{2. Mean Absolute Error (MAE)}
 \[
            \text{J}(\boldsymbol{\theta}) = \frac{1}{N} \sum_{i=1}^{N} |y^{(i)} - f(X^{(i)};\theta)| := \frac{1}{N} \sum_{i=1}^{N} j_i(\theta)
            \]

            In MSE, we need to calculate the squared term but here in MAE, we calculate the absolute of difference between target and predicted values. It is easier to compute in the sense that we do not need to calculate squared now but it is not a smooth function. Furthermore, it is not differentiable at \( y^{(i)} = f(X^{(i)};\theta) \).

\begin{figure}[H]
                \centering
                \includegraphics[width=0.9\textwidth]{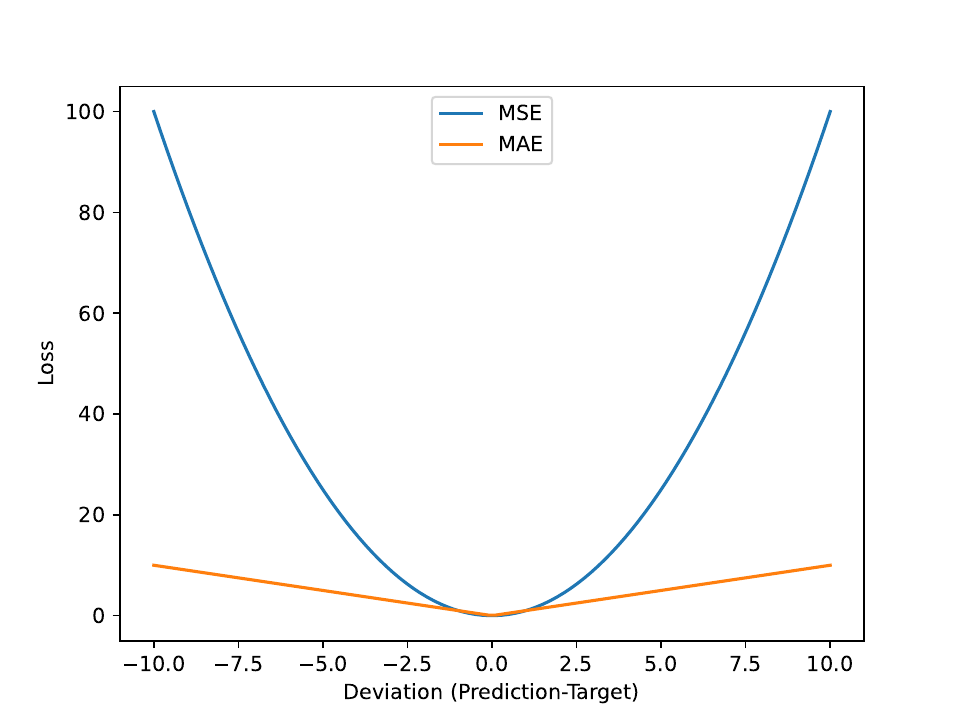}
                \caption{MSE vs MAE}
                \label{fig:msevsmae}
            \end{figure}

\subsubsection{Gradient Descent Variants based on Batch Size}

\textbf{1. Batch Gradient Descent}\\
Batch gradient descent, also known as vanilla gradient descent, computes the gradient of the loss function for the parameters \(\theta\) for the entire training dataset. The following algorithm \ref{alg:bgd} shows how it performs parameter updates.

\begin{algorithm}[H]
\caption{Batch Gradient Descent Algorithm}
\label{alg:bgd}
\begin{algorithmic}[1]
    \State Initialize parameters \( \boldsymbol{\theta} \), learning rate (step length) \(\eta\)
    \While{stopping criterion not met}
        \State Compute gradient estimate: \( g = \frac{1}{N} \sum_{i=1}^{N} \frac{\partial j_i(\theta)}{\partial \boldsymbol{\theta}}  \)
        \State Update parameters: \( \boldsymbol{\theta} \leftarrow \boldsymbol{\theta} - \eta g \)
    \EndWhile
\end{algorithmic}
\end{algorithm}

As we need to calculate the gradients for the whole dataset to perform just one update, batch gradient descent can be very slow and is intractable for datasets that don't fit in memory. Batch gradient descent also doesn't allow us to update our model online, i.e. with new examples on the fly. Batch gradient descent is guaranteed to converge to the global minimum for convex error surfaces and a local minimum for non-convex surfaces. \cite{ruder2017overview}

\vspace{1em}

\textbf{2. Stochastic Gradient Descent}\\
Stochastic gradient descent (SGD) performs a parameter update for each training example \((x^{(i)}, y^{(i)})\). The following algorithm \ref{alg:sgd} shows how it performs parameter updates.

\begin{algorithm}[H]
\caption{Stochastic Gradient Descent Algorithm}
\label{alg:sgd}
\begin{algorithmic}[1]
    \State Initialize parameters \( \boldsymbol{\theta} \), learning rate (step length) \(\eta\)
    \While{stopping criterion not met}
        \State Randomly choose single data point \( (\mathbf{X}^{(i)}, y^{(i)}) \) from \( (\mathbf{X}, \mathbf{y}) \)
        \State Compute gradient estimate: \( g= \frac{\partial}{\partial \theta} j_i(\theta) \)
        \State Update parameters: \( \boldsymbol{\theta} \leftarrow \boldsymbol{\theta} - \eta g \)
    \EndWhile
\end{algorithmic}
\end{algorithm}

While batch gradient descent performs redundant computations for large datasets with similar examples, SGD does parameter updates one at a time. Hence it can be used to learn online.
\vspace{1em}

\textbf{3. Mini-Batch Gradient Descent}
Mini-batch gradient descent takes the best of both batch and stochastic gradient descent and performs a parameter update for every mini-batch of B training examples. The following algorithm \ref{alg:mbgd} shows how it performs parameter updates.

\begin{algorithm}[H]
\caption{Mini-Batch Gradient Descent Algorithm}
\label{alg:mbgd}
\begin{algorithmic}[1]
    \State Initialize parameters \( \boldsymbol{\theta} \), learning rate (step length) \(\eta\), and \textbf{B} batch size.
    \While{stopping criterion not met}
        \State Randomly sample \(B\) examples.
        
        \State Compute gradient estimate: \( g = \frac{1}{B} \sum_{b=1}^{B} \frac{\partial}{\partial \boldsymbol{\theta}} j_b(\theta) \)
        \State Update parameters: \( \boldsymbol{\theta} \leftarrow \boldsymbol{\theta} - \eta g \)
    \EndWhile
\end{algorithmic}
\end{algorithm}

\vspace{1em}

\textbf{Remarks On Gradient Descent Varients Based on Batch Size}
\begin{itemize}
    \item As we are updating parameters after calculating only one gradient in stochastic gradient descent, it will be faster and need little memory to hold gradients, but updating too frequently causes gradients to be noisy and computationally inefficient. But this can also help escape the local minimum and find the global one.
    \item Using batch gradient descent, we update parameters based on gradients of all data points so loading large datasets is problematic. While it could generate smoother losses, calculating the gradients for all data points is time-consuming.
    \item Using mini-batch gradient descent, we update parameters based on a small batch of the data points which will be smoother than stochastic gradient descent and faster than batch gradient descent.
\end{itemize}
It is also worth noting that if we want to perform parameter updates for 1000 iterations, using a batch size of 1, the gradient calculation would be done for only 1000 samples, using a batch size of 32, the gradient calculation would be done for $32*1000$ samples, and using a batch size of 900, the gradient calculation would be done for $900*1000$ samples.

\vspace{1em}

\subsubsection{Gradient Descent Variants based on Parameter Update  \cite{ruder2017overview}}
These techniques often called optimizers in machine learning come in different variations each with their pros and cons. The simplest parameter update techniques are already given in the previous sections in algorithm \ref{alg:simple_gd}. Hence we can start by pointing out its pros and cons. Even though it is easier to calculate and faster, simple gradient descent often oscillates around the local minima. Furthermore, it updates parameters without the knowledge of its past gradient information. To overcome this nature, a momentum optimizer \cite{QIAN1999145} was found.

\vspace{1em}
1. \textbf{Momentum Optimizer}\\
Momentum is a method that helps accelerate SGD in the relevant direction and dampens oscillations. To achieve this, this algorithm adds a fraction $\gamma$ of the previous timestep's update value to the current update value.

Let \(g_{t,k}\) be the gradient of a parameter $k$ with respect to Loss $J(\theta)$ at time step $t$, then we can write it as follows:
\begin{align}
g_{t,k} = \nabla_{\theta_k}J(\theta_{t}) \notag
\end{align}
The momentum term for the current timestep of parameter $k$ is calculated as:
\begin{align}
v_{t,k} &= \gamma v_{t-1,k} +  \eta g_{t,k}
\end{align}
Then the parameter update rule becomes:
\begin{align}
\theta_{t+1,k} &= \theta_{t,k} - v_{t,k}
\end{align}

Where,
\begin{itemize}
    \item  \(v_{t-1}\) is the update vector of the previous time step and is initialized as a zero vector at $t=0$.
    \item \(\gamma\) is a momentum coefficient ranging from 0 to 1 and is usually set to 0.9.
    \item When $\gamma$ is near $0$, it becomes similar to gradient descent but when it is near $1$, helps in faster convergence and reduces oscillations.
\end{itemize}

\break

2. \textbf{Nesterov accelerated gradient}\\
In momentum optimizer, we move our parameters based on the momentum term without caring about where it is heading towards. Now Nesterov Accelerated Gradient (NAG)  calculates the gradient not w.r.t. to our current parameters  \(\theta_t\) but w.r.t. the approximate future position of our parameters $\theta_{t}$. 
For a $k^{th}$ parameter at time step $t$, we perform parameter update as follows:

\begin{align} \label{eq:nag}
\begin{split} 
v_{t,k} &= \gamma v_{t-1,k} + \eta \nabla_{\theta_k} J( \theta_{t} - \gamma v_{t-1} ) \\ 
\theta_{t+1,k} &= \theta_{t,k} - v_{t,k} 
\end{split} 
\end{align}

Here $\gamma$, a momentum term is set around 0.9 again.
\begin{figure}[H]
    \centering
    \includegraphics[width=0.3\textwidth]{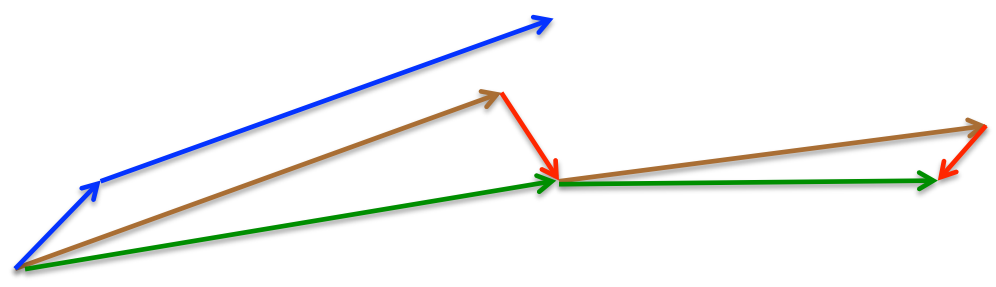}
    \caption{Nesterov update (Source: G. Hinton’s lecture 6c) }
    \label{fig:nag}
\end{figure}
While Momentum first computes the current gradient (small blue vector in the above Figure) and then takes a big jump in the direction of the updated accumulated gradient (big blue vector), NAG first makes a big jump in the direction of the previously accumulated gradient (brown vector), measures the gradient and then corrects (green vector).

\vspace{1em}

3. \textbf{Adaptive Gradient Method (Adagrad) \cite{duchi2011adaptive}}\\
Let \(g_{t,k}\) be the gradient of a parameter $k$ with respect to Loss $J(\theta)$ at time step $t$, then we can write it as follows:
\begin{align}
    g_{t, k} = \nabla_{\theta_k} J( \theta_{t} ) \notag
\end{align}

In previous methods, we had a constant learning rate for all parameters. However, this is problematic for frequent and infrequent parameters. Adagrad adapts the learning rate to the parameters, performing larger updates for infrequent and smaller updates for frequent parameters. 

Let $G_t$ be the vector to contain the sum of squares of gradients and can be written for parameter $k$ as:

\begin{align}
    G_{t,k} = G_{t-1,k}+g_{t,k}^2 \notag
\end{align}

Then the update rule is:
\begin{align}
\theta_{t+1, k} = \theta_{t, k} - \dfrac{\eta}{\sqrt{G_{t,k} + \varepsilon}} \cdot g_{t, k}
\end{align}
Here, $\varepsilon$ is a smoothing term that avoids division by zero (usually on the order of $1^{-8}$ ). $G_t$ at $t=1$ is a zero vector. Now we have an adaptive learning rate, but as we accumulate squared gradients in the denominator, the adapted learning rate can shrink quickly.

\vspace{1em}

4. \textbf{RMSprop \cite{tieleman2012lecture}}\\

Instead of the sum of squares of the gradient, we now use the running average of the squares of the gradients. For parameter $\theta_k$ at time step $t$, we can define the running average of the square of the gradients as:
\begin{align}
E[g^2]_{t,k} = \gamma E[g^2]_{t-1,k} + (1 - \gamma) g^2_{t,k} \notag
\end{align}

The parameter update rule is:

\begin{align} 
\theta_{t+1,k} = \theta_{t,k}- \dfrac{\eta}{\sqrt{E[g^2]_{t,k} + \epsilon}} g_{t,k}  :=  \theta_{t,k}- \dfrac{\eta}{RMS[g]_{t,k}} g_{t,k} \notag
\end{align}

Where RMS is the root mean squared criterion of the gradient. Hence the name \textbf{RMS}. The author (G. Hinton) suggested using \(\gamma=0.9\) and \(\eta=0.001\). 

\vspace{1em}

5. \textbf{ADADELTA \cite{zeiler2012adadelta}} \\
The authors\cite{zeiler2012adadelta}, mention that this optimizer was derived to improve the two main drawbacks of Adagrad, 1) the continual decay of learning rates
throughout training, and 2) the need for a manually selected global learning rate.

Although being developed independently around the same time, RMSProp and Adadelta work similarly around the RMS term. Authors first derived the update rule as:

\begin{align} 
\theta_{t+1,k} =  \theta_{t,k}- \dfrac{\eta}{RMS[g]_{t,k}} g_{t,k} \notag
\end{align}

But authors \cite{zeiler2012adadelta} note that units in these steps do not match (i.e. if the parameters had some hypothetical units, the changes to the parameter should be changes in those units as well) so they defined \textbf{running average of squared parameter updates}:
\begin{align}
E[\Delta \theta^2]_{t,k}=\gamma E[\Delta \theta^2]_{t-1,k} + (1 - \gamma) \Delta \theta^2_{t,k} \notag
\end{align}

And its RMS is, 
\begin{align}
RMS[\Delta \theta]_{t,k} = \sqrt{E[\Delta \theta^2]_{t,k} + \epsilon}    \notag
\end{align}

Since \(RMS[\Delta \theta]_{t}\) is unknown, we approximate it with the RMS of parameter updates until the previous timestep to get a new parameter update rule:

\begin{align}
\theta_{t+1,k} = \theta_{t,k} - \eta \dfrac{RMS[\Delta \theta]_{t-1,k}}{RMS[g]_{t,k}} g_{t,k}
\end{align}

The original implementation does not have any learning rate i.e. $\eta=1$.

\vspace{1em}

6. \textbf{ADAM (Adaptive Moment Estimation) \cite{kingma2014adam}}\\

Adam combines RMSProp's exponentially decaying average of past squared gradients (\(v_t\)) and exponentially decaying average of past gradients (like momentum, \(m_t\)). For parameter $\theta_k$ at time $t$:
\begin{align}
m_{t,k} = \beta_1 m_{t-1,k} + (1 - \beta_1) g_{t,k} \notag \\
v_{t,k} = \beta_2 v_{t-1,k} + (1 - \beta_2) g_{t,k}^2 \notag
\end{align}

Here, $\beta_1$ and $\beta_2$ are called decay rates. 

As \(m_t\) (first moment of gradients) and \(v_t\) (second moment of gradients) are initialized as vectors of zeros, they tend to bias towards 0 in early timesteps when decay rates are low. Hence authors proposed bias-corrected moment estimates.
\begin{align}
\hat{m}_{t,k} = \frac{m_{t,k}}{1 - \beta_1^t} \notag \\
\hat{v}_{t,k} = \frac{v_{t,k}}{1 - \beta_2^t} \notag
\end{align}

Then the parameter update rule becomes:
\begin{align}
    \theta_{t+1,k} = \theta_{t,k} - \frac{\eta}{\sqrt{\hat{v}_{t,k}} + \varepsilon} \hat{m}_{t,k} \label{eq:adam_update}
\end{align}

The authors propose default values of 0.9 for \(\beta_1\), 0.999 for \(\beta_2\), and \(10^{-8}\) for \(\varepsilon\).

\break

\section{Experiments and Results}
In this section, above mentioned optimizers are experimented on the toy dataset of the least squares problem. The dataset is prepared pseudo-randomly and has the following properties:
\begin{itemize}
    \item Random Seed: \(100\)
    \item Input Data: \( \mathbf{X} \in \mathbb{R}^{1000 \times 5} \)
    \item Random data \( \mathbf{X} \sim \mathcal{U}(0, 100) \)
    \item Normalize data by dividing by max value: \( \mathbf{X}_{\text{norm}} = \frac{\mathbf{X}}{\max(\mathbf{X})} \)
    \item Parameters (or Weights): \( \boldsymbol{\theta} \sim \mathcal{N}(0, 1) \), \( \boldsymbol{\theta} \in \mathbb{R}^{5} \)
    \item Bias: \( b \sim \mathcal{N}(0, 1) \), \( b \in \mathbb{R} \)
    \item Noise: \( \epsilon \sim \mathcal{N}(0, 1) \), \( \epsilon \in \mathbb{R}^{1000} \)
    \item Output: \( \mathbf{y} = \mathbf{X}_{\text{norm}} \boldsymbol{\theta} + b + 0.1\epsilon \), \( \mathbf{y} \in \mathbb{R}^{1000} \)
\end{itemize}
\vspace{1em}

Our goal will be to find the parameters $\theta$ which will be able to minimize the loss function.
\vspace{1em}

To evaluate the performance of the parameter in each epoch, we randomly split \(X_{norm}\) into \(90:10\) ratios. The first half will be used to calculate loss and gradients to update parameters while the second half will be used to evaluate the parameter's performance on unseen data. The first half is often called \textbf{training data} and the second half is called \textbf{validation data}.\\

\subsection{Experiments Setup}
For this experiment, we can have:
\begin{itemize}
    \item variations of loss function as 1 i.e. MSE
    \item variations of batch size: 1, 32, whole dataset
    \item variations of learning rates (if applicable): 0.1, 0.01, 0.001, and for momentum: 0.1, 0.9
    \item variations of optimizers are 7 (SGD, Momentum, Nesterov, Adagrad, RMSProp, Adadelta, Adam)
    \item number of epochs (i.e. number of iterations) as 1000.
\end{itemize}
Experiments are done with the following settings:
\begin{itemize}
    \item PyTorch is used for training a linear model. Because it handles gradient calculations and provides optimizer's implementations.
    \item Parameters in the linear model are initialized $\mathcal{U}(-\frac{1}{\sqrt{5}}, \frac{1}{\sqrt{5}})$
\end{itemize}
\break
\textbf{Note:} PyTorch implementation of Momentum and NAG optimizers are slightly different than the proposed by \fullfootcite{sutskever2013importance}. But the PyTorch's implementation has been used in some popular works like DenseNet \fullfootcite{huang2018denselyconnectedconvolutionalnetworks}. In previous slides, we showed implementation by \fullfootcite{ruder2017overview}.

\begin{table}[h!]
    \centering
    \begin{tabular}{|l|p{6cm}|p{6cm}|}
        \hline
        & \textbf{\cite{sutskever2013importance} Implementation} & \textbf{PyTorch Implementation} \\
        \hline
        \textbf{Momentum} & 
        \( v_{t} = \mu v_{t-1} - \eta \nabla_\theta J(\theta_{t-1}) \) \par
        \( \theta_{t} = \theta_{t-1} + v_{t} \) &
        \( v_{t} = \mu v_{t-1} + \nabla_\theta J(\theta_{t-1}) \) \par
        \( \theta_{t} = \theta_{t-1} - \eta v_{t} \) \\
        \hline
        \textbf{NAG} & 
        \( v_{t} = \mu v_{t-1} - \eta \nabla_\theta J(\theta_{t-1} + \mu v_{t-1}) \) \par
        \( \theta_{t} = \theta_{t-1} + v_{t} \) &
        \( v_{t} = \mu v_{t-1} + \nabla_\theta J(\theta_{t-1}) \) \par
        $g_t = g_t + \mu v_t$ \par
        \( \theta_{t} = \theta_{t-1} - \eta g_{t} \) \\
        \hline
    \end{tabular}
    \caption{Comparison of Original and PyTorch Implementations of Momentum and NAG}
\end{table}

\subsection{Results}
Based on the learning rates, we can group optimizers into two groups i.e. constant learning rate optimizers and adaptive learning rate optimizers.

\subsubsection{Validation Loss On Constant LR Optimizers}
We have SGD, Momentum, and Nesterov adaptive gradient for the constant learning rate-based optimizers.
\vspace{1em}

1. \textbf{Stochastic Gradient Descent Optimizer}
\begin{center}
\begin{figure}[H]
    \includegraphics[width=1\textwidth]{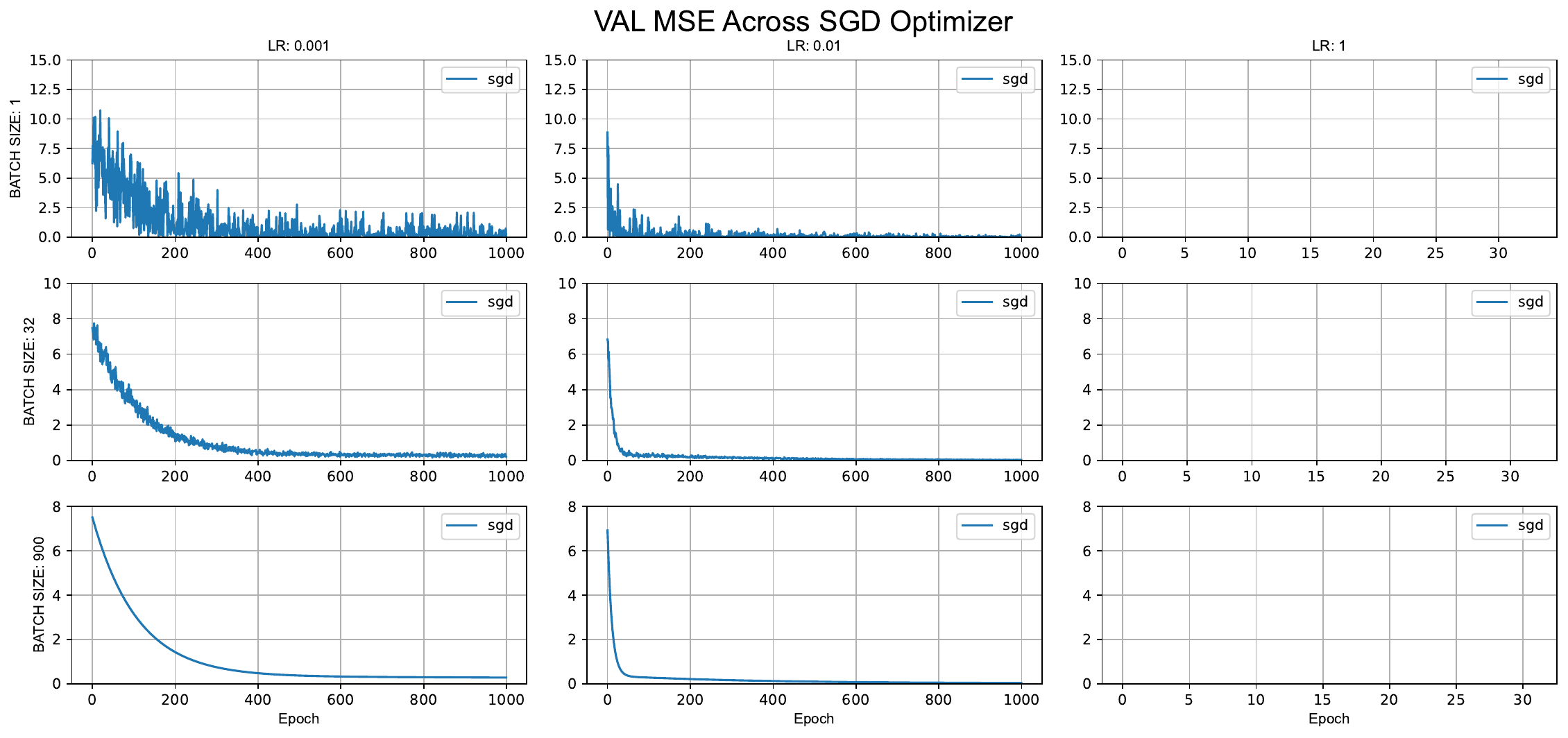} 
    \caption{Validation MSE While using SGD}
    \label{sgd_plot} 
\end{figure}
\end{center}

From figure \ref{sgd_plot} we can observe that:
\begin{itemize}
    \item Using a batch size of 1 (i.e. stochastic gradient descent), the loss decreases with too much noise i.e. unstable. It is an expected behavior when we update parameters based on the gradient of only one training example. By the end of 1000 iterations, we calculated gradients for only 1000 samples.
    \item Using a batch size of 32 (i.e. mini-batch gradient descent), the loss decreases slightly more than the batch size of 1. This is because, in each epoch, we update parameters based on the gradients of 32 training examples. By the end of 1000 iterations, we calculated gradients for 32*1000 samples.
    \item Using a batch size of 900 (i.e. batch gradient descent), loss values are much smoother and do not oscillate either. It is because in each epoch we are updating parameters based on the gradients of 900 training examples.  By the end of 1000 iterations, we calculated gradients for 900*1000 samples.
    \item As the learning rate increases, we are experiencing less loss quickly but at learning rate 1, everything diverges. It happened because the gradients were too large and hence parameters were i.e. gradient explosion.
\end{itemize}

2. \textbf{Momentum Optimizer}
\begin{center}
\begin{figure}[H]
    \includegraphics[width=1\textwidth]{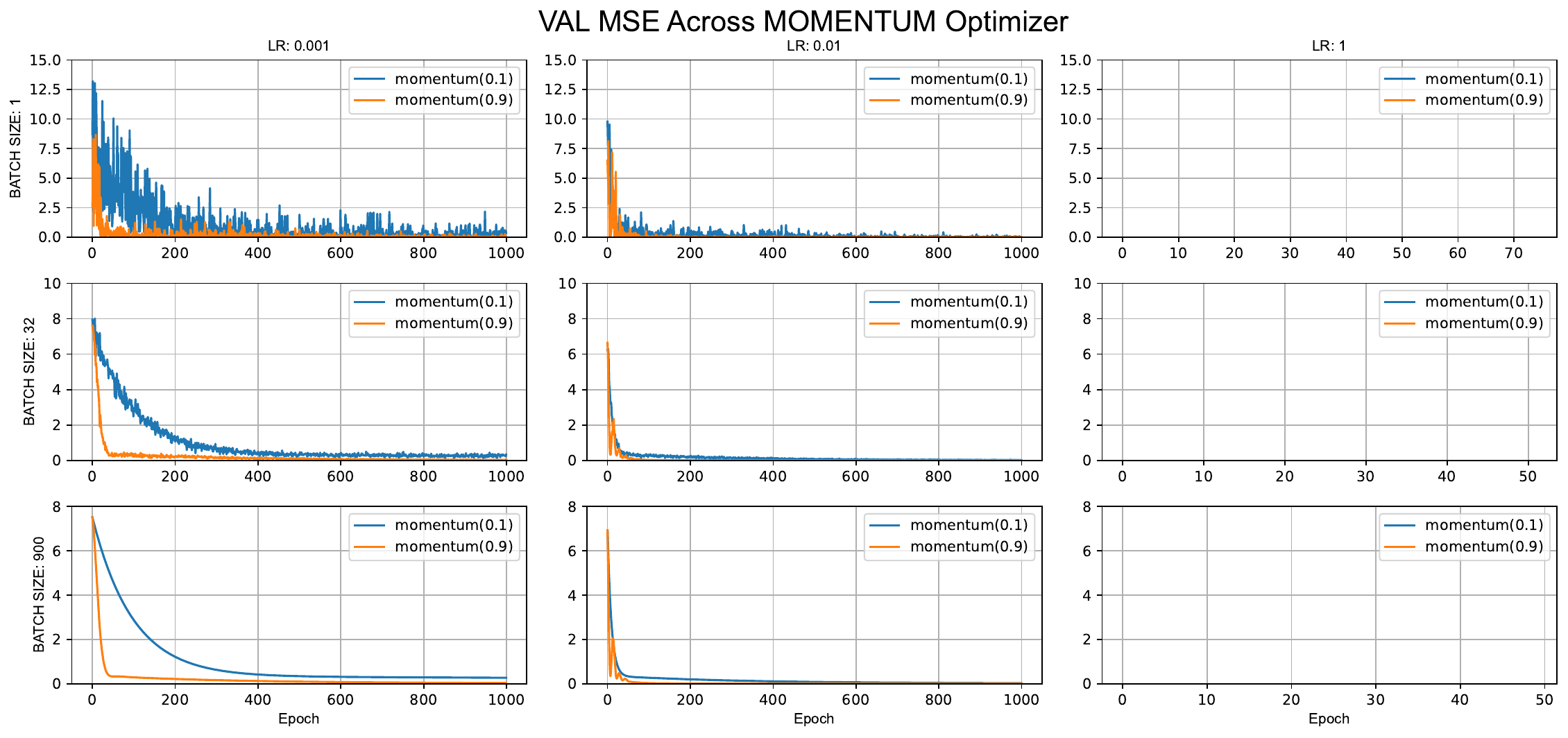} 
    \caption{Validation MSE While using Momentum Optimizer}
    \label{momentum_plot} 
\end{figure}
\end{center}

From figure \ref{momentum_plot} we can observe that:
\begin{itemize}
    \item Loss curves follow a pattern like when using SGD. i.e. as batch size increases loss is less noisy and at learning rate 1, the loss was too high and training was aborted.  
    \item When the momentum rate is too low, it behaves more like SGD, but when it is high we see more stable and smaller losses. 
\end{itemize}

3. \textbf{Nesterov Optimizer}
\begin{center}
\begin{figure}[H]
    \includegraphics[width=1\textwidth]{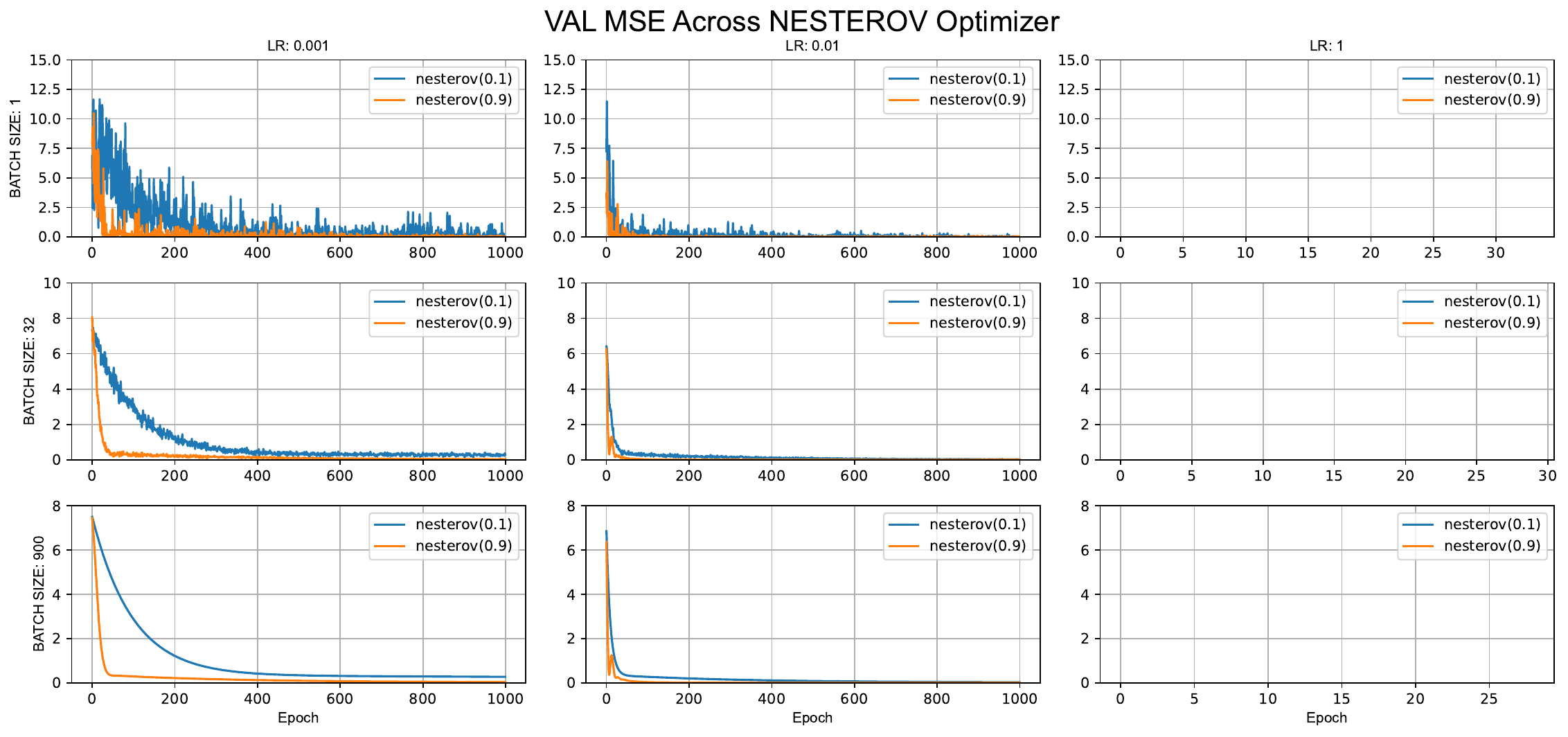} 
    \caption{Validation MSE While using Nesterov Optimizer}
    \label{nesterov_plot} 
\end{figure}
\end{center}

From figure \ref{nesterov_plot} we can observe that:
\begin{itemize}
    \item Loss curves follow a pattern like a momentum optimizer. i.e. as batch size, momentum, and rate increases loss is less noisy, and at learning rate 1, the loss was too high and training was aborted.
\end{itemize}

Finally, we can compare them in a single plot.

\begin{center}
\begin{figure}[H]
    \includegraphics[width=1\textwidth]{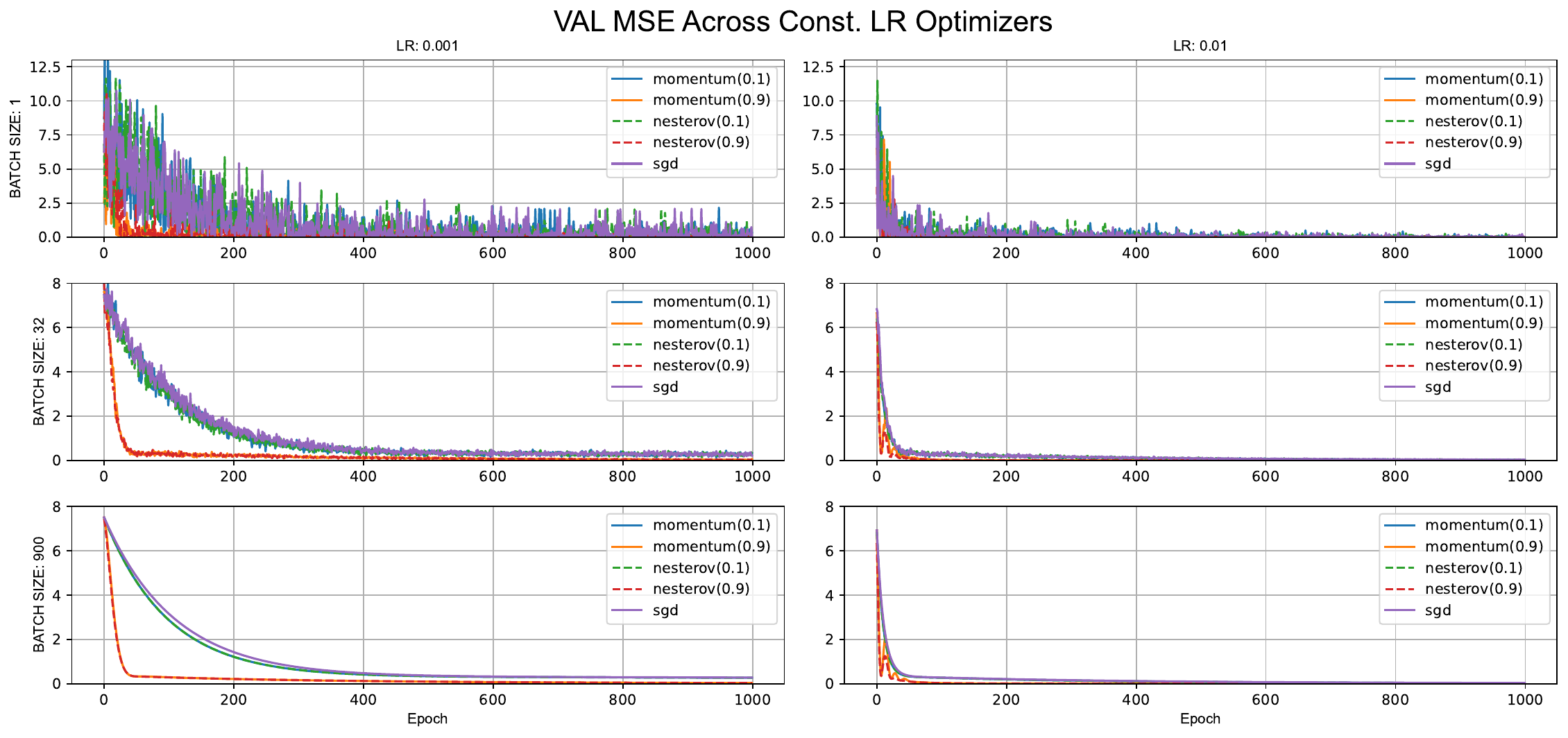} 
    \caption{Validation MSE across constant LR optimizers}
    \label{const_lr_loss} 
\end{figure}
\end{center}
From figure \ref{const_lr_loss} we can observe that:
\begin{itemize}
    \item Nesterov and Momentum both have almost similar loss curves for this data.
    \item Using Nesterov, we can see slightly smoother loss curves than momentum with a batch size of 32 and batch size of 900.
\end{itemize}

\subsubsection{Validation Loss On Adaptive LR Optimizers}
We have Adagrad,  RMSProp, Adadelta, and Adam for the adaptive learning rate-based optimizers.

1. \textbf{Adagrad Optimizer}
\begin{center}
\begin{figure}[H]
    \includegraphics[width=1\textwidth]{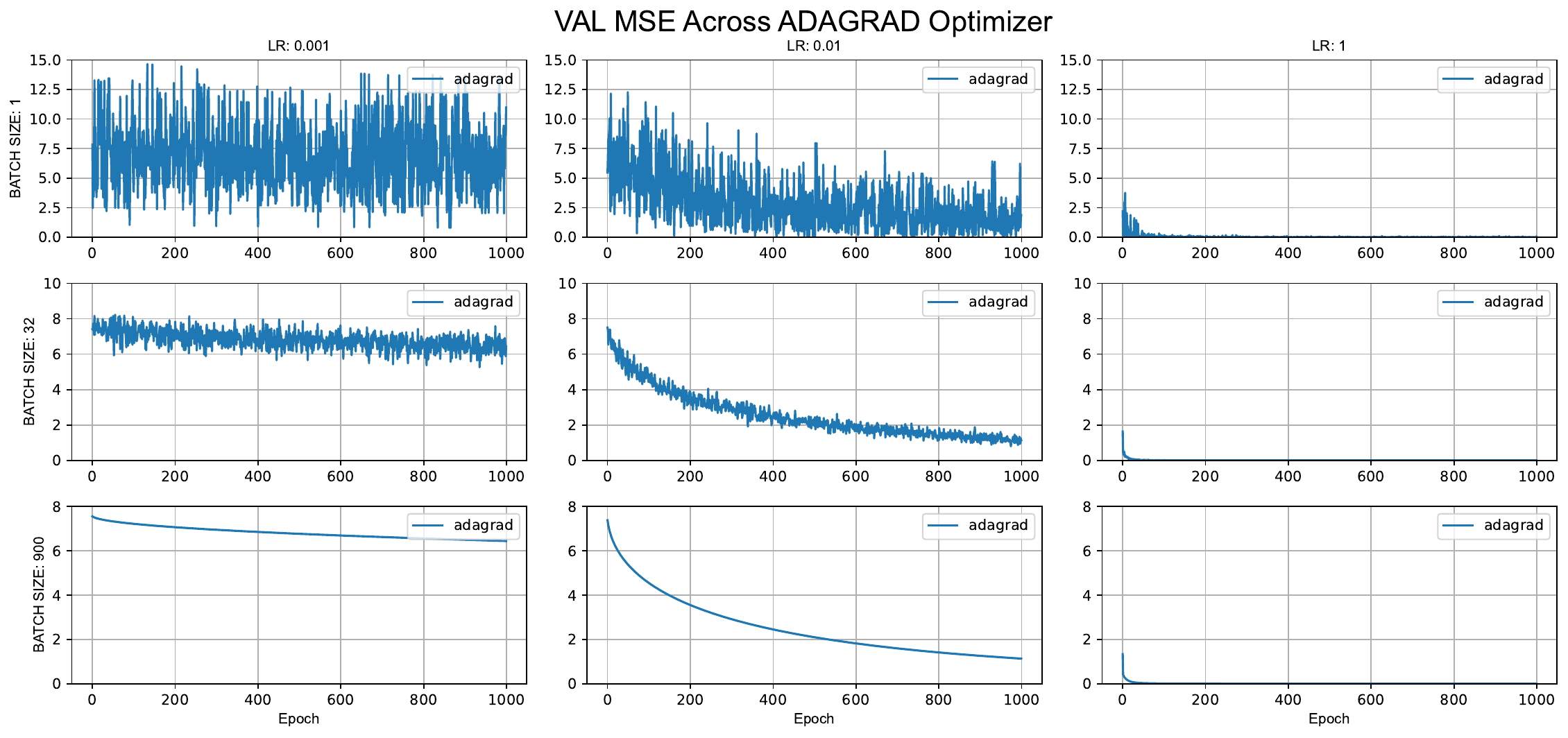} 
    \caption{Validation MSE While using Adagrad Optimizer}
    \label{adagrad_plot} 
\end{figure}
\end{center}

From figure \ref{adagrad_plot} we can observe that:
\begin{itemize}
\item It seems higher learning rates have reduced loss more and now gradient explosion also did not happen at a learning rate of 1. It is because we do not have a fixed learning rate now but it is being adapted and different for different parameters.
\end{itemize}

2. \textbf{RMSProp Optimizer}
\begin{center}
\begin{figure}[H]
    \includegraphics[width=1\textwidth]{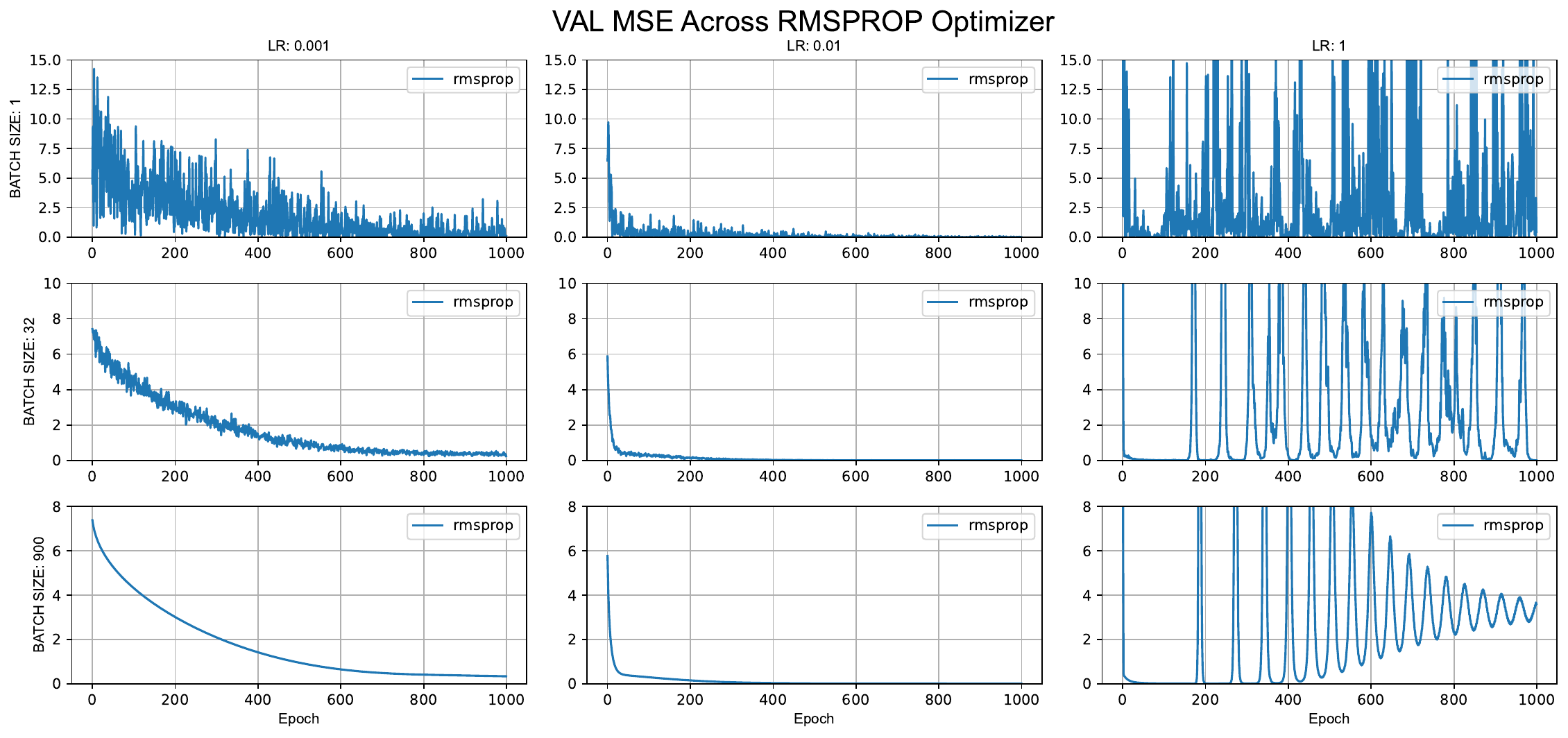} 
    \caption{Validation MSE While using RMSProp Optimizer}
    \label{rmsprop_plot} 
\end{figure}
\end{center}

From figure \ref{rmsprop_plot} we can observe that:
\begin{itemize}
\item A learning rate of 1 gives giving very small loss in the early epoch but it shows very unstable loss behavior.
\end{itemize}

3. \textbf{Adadelta Optimizer}
\begin{center}
\begin{figure}[H]
    \includegraphics[width=1\textwidth]{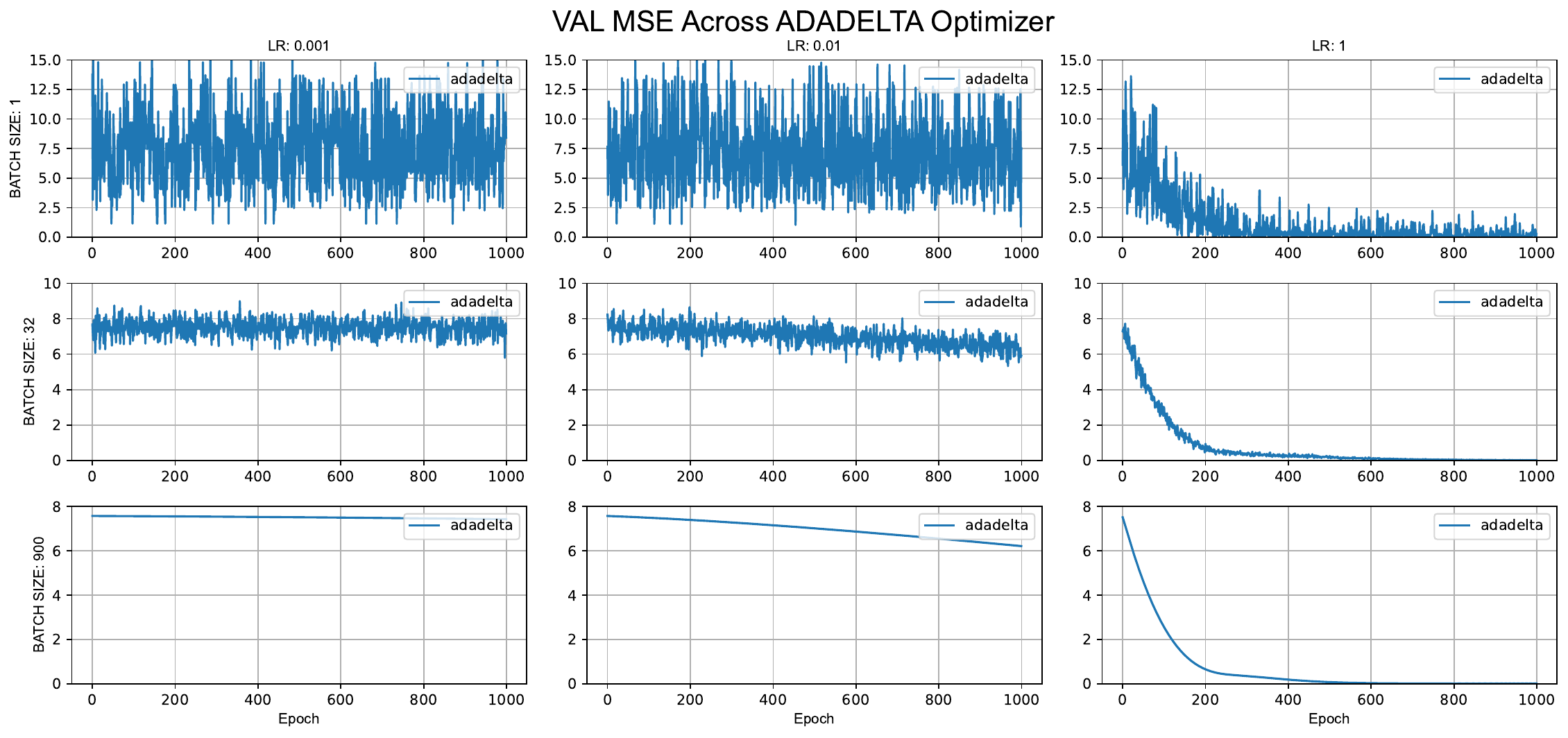} 
    \caption{Validation MSE While using Adadelta Optimizer}
    \label{adadelta_plot} 
\end{figure}
\end{center}

From figure \ref{adadelta_plot} we can observe that:
\begin{itemize}
\item A learning rate of 1 is giving smoother and better results than others. This is expected because authors \cite{zeiler2012adadelta} used a learning rate of 1.
\end{itemize}

4. \textbf{Adam Optimizer}
\begin{center}
\begin{figure}[H]
    \includegraphics[width=1\textwidth]{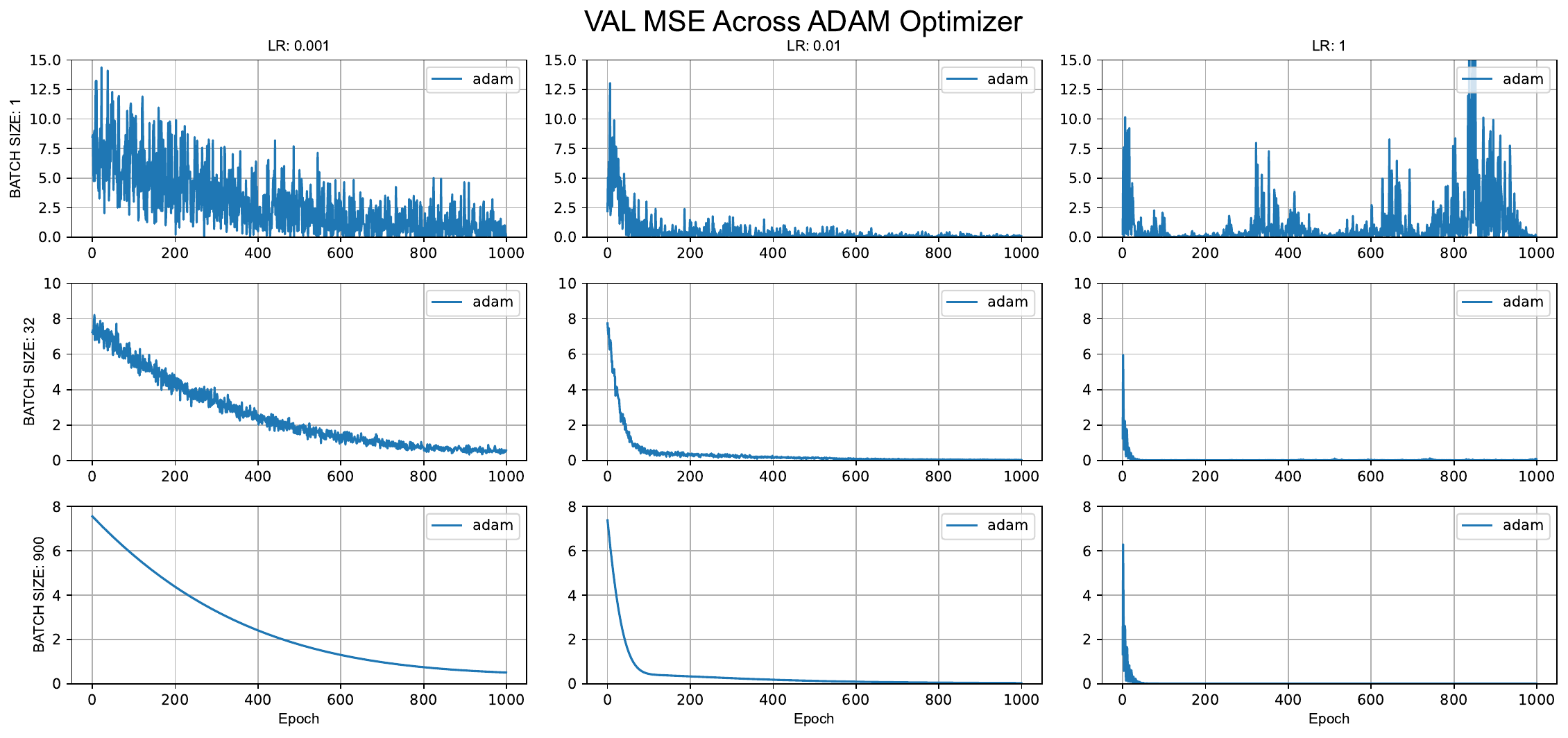} 
    \caption{Validation MSE While using Adam Optimizer}
    \label{adam_plot} 
\end{figure}
\end{center}

From figure \ref{adam_plot} we can observe that:
\begin{itemize}
\item At a learning rate of 1 and batch size of 1, Adam is unstable but increasing batch size shows stable behavior.
\end{itemize}

Now we can look at all these 4 results in a single plot to make comparisons.

\begin{center}
\begin{figure}[H]
    \includegraphics[width=0.85\textwidth]{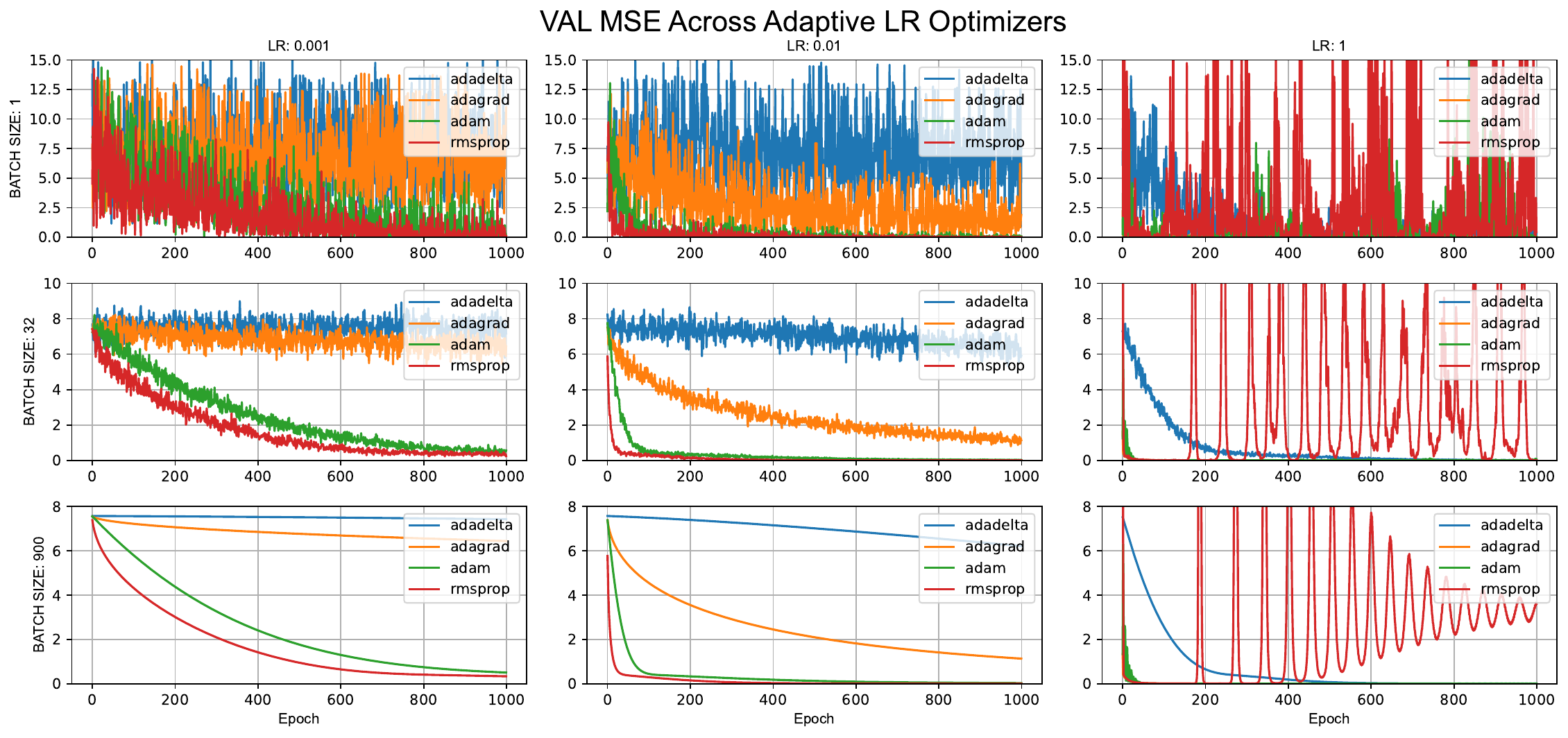} 
    \caption{Validation MSE across adaptive LR optimizers}
    \label{var_lr_loss} 
\end{figure}
\end{center}

From figure \ref{var_lr_loss} we can observe that,
\begin{itemize}
    \item Only when the learning rate is 1, is the loss reduced faster for Adadelta.
    \item Other than RMSProp, optimizers quickly reached lesser loss values at a learning rate of 1.
    \item Adam and RMSProp quickly achieved a smaller validation MSE than others but RMSProp shows instability at a learning rate of 1.
\end{itemize}

We can look into the Nesterov and Adam optimizers for comparison as well because these two are the better-performing optimizers.

\begin{center}
\begin{figure}[H]
    \includegraphics[width=0.85\textwidth]{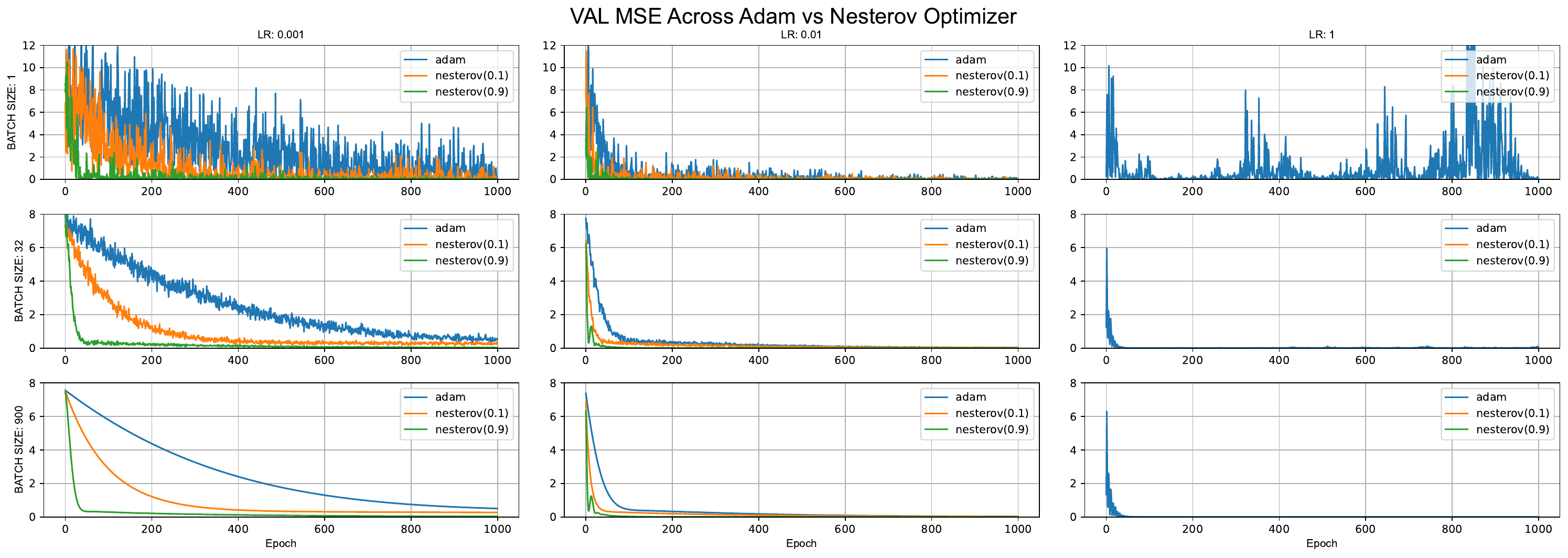} 
    \caption{Validation MSE across Adam and Nesterov optimizers}
    \label{adam_nesterov} 
\end{figure}
\end{center}

From figure \ref{adam_nesterov} we can observe that,
\begin{itemize}
    \item Nesterov is performing better than Adam in almost all of the experiments.
    \item While Nesterov failed in learning rate 1, Adam has reached a smaller loss quickly.
\end{itemize}

\section{Discussion}
Based on the above experiments, we can conclude that:
\begin{itemize}
    \item Using mini-batch gradient descent we can leverage the properties of SGD (faster update but higher noisy gradients) and Full GD (slower update but smoother gradients). i.e. tradeoff between faster updates and smoother gradients.
    \item Using a higher learning rate can take a bigger update step and might reach minimum loss faster but can cause gradient explosion as well. Using a lower learning rate takes a smaller update step but needs more iterations to reach minimum loss.
    \item Using adaptive optimizers, Adam performed better than others.
    \item Using fixed learning rate-based optimizers, Nesterov's performance was observed to be the best.
    \item When working on large-scale datasets, adaptive learning rate-based optimizers are best as they update parameters with different learning rates or step lengths in each iteration for individual parameters. But can be slower due to requiring additional computations.    
\end{itemize}

For our experiment, noise was selected to be small and it would be interesting to see how well loss values decrease upon increasing the noise level.

\printbibliography

\end{document}